\documentclass[12pt]{amsart}
\usepackage{amsmath}
\usepackage{amsfonts,amssymb,amsthm}
\usepackage{graphicx}

\newcommand{\beq}{\begin{equation}}
\newcommand{\eeq}{\end{equation}}
\newcommand{\bbar}{\begin{eqnarray}}
\newcommand{\eear}{\end{eqnarray}}

\newcommand{\thm}[2]{\begin{#1} #2 \end{#1}}

\newtheorem{theorem}{Theorem}[section]
\newtheorem{itheorem}{Theorem}[section]
\newtheorem{lemma}[theorem]{Lemma}

\begin{document}

\title{Yet another zeta function and learning}

\author{Igor Rivin}
\address{Mathematics department, University of Manchester,
Oxford Road, Manchester M13 9PL, UK}
\address{Mathematics Department, Temple University,
Philadelphia, PA 19122}
\address{Mathematics Department, Princeton University, Princeton,
NJ 08544}
\email{irivin@math.princeton.edu} \thanks{The author would like 
to think the EPSRC and the NSF for support, and Natalia Komarova 
and  Ilan Vardi for useful conversations. }

\begin{abstract}
We analyze completely the convergence speed of the \emph{batch 
learning algorithm}, and compare its speed to that of the 
memoryless learning algorithm and of learning with memory (as 
analyzed in \cite{kr2}). We show that the batch learning 
algorithm is never worse than the memoryless learning algorithm 
(at least asymptotically). Its performance \emph{vis-a-vis} 
learning with full memory is less clearcut, and depends on 
certain probabilistic assumptions. These results necessitate the 
introduction of the \textit{moment zeta function} of a 
probability distribution and the study of some of its properties. 
\end{abstract}
\maketitle

\renewcommand{\theitheorem}{\Alph{itheorem}}
\section*{Introduction}
The original motivation for the work in this paper was provided 
by  research in learning theory, specifically in various models 
of language acquisition (see, for example, \cite{knn,nkn,kn}). In 
the paper \cite{kr2}, we had studied the speed of convergence of 
the  \emph{memoryless learner algorithm}, and also of 
\emph{learning with full memory}. Since the \emph{batch learning 
algorithm} is both widely known, and believed to have superior 
speed (at the cost of memory) to both of the above methods by 
learning theorists, it seemed natural to analyze its behavior 
under the same set of assumptions, in order to bring the analysis 
in \cite{kr1} and \cite{kr2} to a sort of closure. It should be 
noted that the detailed analysis of the batch learning algorithm 
is performed under the assumption of \emph{independence}, which 
was not explicitly present in our previous work. For the 
impatient reader we state our main result (Theorem 
\ref{batchthm}) immediately (the reader can compare it with the 
results on the memoryless learning algorithm and learning with 
full memory, as summarized in Theorem \ref{mainprev}): 
\begin{itheorem}
 Let $N_\Delta$ be the number of steps it takes 
for the student (with probability $1$) to have probability $1 - 
\Delta$ of learning the concept using the batch learner 
algorithm. Then we have the following estimates for $N_\Delta$:
\begin{itemize}
\item
If the distribution of overlaps is \emph{uniform}, or more 
generally, the density function $f(1-x)$  at $0$ has the form 
$f(x) = c + O(x^\delta),$ $\delta, c > 0,$ then $N_\Delta=|\log 
\Delta|\Theta(n)$ 
\item 
If the probability density function $f(1-x)$ is asymptotic to 
$x^\beta + O(x^{\beta - \delta}), \quad \delta, \beta > 0$, as 
$x$ approaches $0$, then we have $N_\Delta=|\log 
\Delta|\Theta(n^{1/(1+\beta)})$; 
\item 
If the asymptotic behavior is as above, but $-1/2 < \beta < 0$, 
then $N_\Delta=|\log \Delta|\Theta(n^{1/(1+\beta)}).$ 
\end{itemize}
\end{itheorem}
The plan of the paper is as follows: in this Introduction we 
recall the learning algorithms we study; in Section \ref{mathmod} 
we define our mathematical model; in Section 2 we recall our 
previous results, in Section 3 we begin the analysis of the batch 
learning algorithm, and introduce some of the necessary 
mathematical concepts; in Sections 4-6 we analyze the three cases 
stated in Theorem A, and we summarize our findings in Section 7.
\subsection*{Memoryless Learning and Learning with Full Memory} 
The general setup is as follows: There is a collection of 
concepts $R_0, \dots, R_n$ and words which refer to these 
concepts, sometimes ambiguously. The teacher generates a stream 
of words, referring to the concept $R_0$. This is not known to 
the student, but he must learn by, at each step, guessing some 
concept $R_i$ and checking for consistency with the teacher's 
input.  The \emph{memoryless learner algorithm} consists of 
picking a concept $R_i$ at random, and sticking by this choice, 
until it is proven wrong.  At this point another concept is 
picked randomly, and the procedure repeats. \emph{Learning with 
full memory} follows the same general process with the important 
difference that once a concept is rejected, the student never 
goes back to it. It is clear (for both algorithms) that once the 
student hits on the right answer $R_0$, this will be his final 
answer. We would like to estimate the probability of having 
guessed the right answer is after $k$ steps, and also the 
expected number of steps before the student settles on the right 
answer.

\subsection*{Batch Learning} The batch learning situation is 
similar to the above, but here the student records the words 
$w_1, \dots, w_k, \dots$ he gets from the teacher. For each word 
$w_i$ , we assume that the student can find (in his textbook, for 
example) a list $L_i$ of concepts referred to by the word. If we 
define 
\begin{equation*} 
\mathcal{L}_k = \bigcap_{i=1}^k L_i,
\end{equation*}
then we are interested in the smallest value of $k$ such that 
$\mathcal{L}_k = \{R_0\}$. This value $k_0$ is the time it has 
taken the student to learn the concept $R_0$. We think of $k_0$ 
as a random variable, and we wish to estimate its expectation.
\section{The mathematical model}
\label{mathmod}
 We think of the words referring to the concept 
$R_0$ as a probability space $\mathcal{P}$. The probability that 
one of these words also refer to the concept $R_i$ shall be 
denoted by $p_i$; the probability that a word refers to concepts 
$R_{i_1}, \dots, R_{i_k}$ shall be denoted by $p_{i_1 \dots 
i_k}$. All the results described below (obviously) depend in a 
crucial way on the $p_1, \dots, p_n$ and (in the case of the 
batch learning algorithm) also on the joint probabilities. Since 
there is no \emph{a priori} reason to assume specific values for 
the probabilities, we shall assume that all of the $p_i$ are 
themselves \emph{independent, identically distributed random 
variables}. We shall refer to their common distribution as 
$\mathcal{F}$, and to the density as $f$. It turns out that the 
convergence properties of the various learning algorithms depend 
on the local analytic properties of the distribution 
$\mathcal{F}$ at $1$ -- some moments reflection will convince the 
reader that this is not really so surprising. 

To carry out a precise analysis of the batch learning algorithm, 
we will also need the \emph{independence hypothesis}:
$$
p_{i_1 \dots i_k} = p_{i_1} \dots p_{i_k}.
$$
It is again not too surprising that some such assumption on 
correlations ought to be required for precise asymptotic results, 
though it is obviously the subject of a (non-mathematical) debate 
as to whether assuming that the various concepts are truly 
independent is reasonable from a cognitive science point of view. 

\section{Previous results}
In previous work \cite{kr1} and \cite{kr2} we obtained the 
following result. 
 \thm{theorem} {\label{mainprev}Let $N_\Delta$ be the number of steps it 
takes for the student (with probability $1$) to have probability 
$1 - \Delta$ of learning the concept. Then we have the following 
estimates for $N_\Delta$:
\begin{itemize}
\item
if the distribution of overlaps is \emph{uniform}, or more 
generally, the density function $f(1-x)$  at $0$ has the form 
$f(x) = c + O(x^\delta),$ $\delta, c > 0,$ then 
$N_\Delta=|\log \Delta|\Theta(n \log n)$ for 
the memoryless algorithm and $N_\Delta=(1-\Delta)^2 \Theta(n \log 
n)$ when learning with full memory; 
\item 
if the probability density function $f(1-x)$ is asymptotic to 
$x^\beta + O(x^{\beta - \delta}), \quad \delta, \beta > 0$, as 
$x$ approaches $0$, then for the two algorithms we have 
respectively $N_\Delta=|\log \Delta|\Theta(n)$ and 
$N_\Delta=(1-\Delta)^2 \Theta(n)$;
\item 
if the asymptotic behavior is as above, but $-1 < \beta < 0$, then
$N_\Delta=|\log \Delta|\Theta(n^{1/(1+\beta)})$ for the memoryless
learner and $(N_\Delta=1-\Delta)^2\Theta(n^{1/(1+\beta)})$ for
learning with full memory.
\end{itemize}}
\noindent Recall that $f(x) = \Theta(g(x))$ means that for 
sufficiently large $x$, the ratio $f(x)/g(x)$ is bounded between 
two strictly positive constants. The distribution of overlaps 
referred to above is simply the distribution $\mathcal{F}$. 
Notice that the theorem says nothing about the situation when 
$\mathcal{F}$ is supported in some interval $[0, a]$, for $a<1$. 
That case is (presumably) of scientific interest, but 
mathematically it is relatively trivial: we replace the arguments 
of all the $\Theta$s above by $1$, though, of course, we are 
thereby hiding the dependence on $a$.

\section{General bounds on the batch learner algorithm}

Consider a set of words $w_1, \dots, w_k$. The probability that 
they all refer to the concept $R_i$ is, obviously $p_i^k$. 
\begin{lemma}
\label{bounds}
 The probability $q_k$ that we still have not 
learned the concept $R_0$ after $k$ steps is bounded above by 
$\sum_{i=1}^n p_i^k$, and below by $\max_i p_i^k$. 
\end{lemma} 
\begin{proof}
Immediate. 
\end{proof}
We will first use these upper and lower 
bounds to get corresponding bounds on the convergence speed of 
the batch learner algorithm, and then invoke the independence 
hypothesis to sharpen these bounds in many cases.

We begin with a trivial but useful lemma.
\begin{lemma}
\label{rearrange}
 Let $G$ be a game where the probability of 
success (respectively failure) after at most $k$ steps is $s_k$ 
(respectively $f_k = 1-s_k $). Then the expected number of steps 
until success is 
$$\sum_{k=1}^\infty k (s_k - s_{k-1}) = \sum_{k=1}^\infty s_k = 1 - 
\sum_{k=1}^\infty f_k,$$ if the corresponding sum converges.
\end{lemma}
\begin{proof}
The proof is immediate from the definition of expectation and the 
possibility of rearrangment of terms of positive series.
\end{proof}
We can combine Lemma \ref{rearrange} and Lemma \ref{bounds} to 
obtain:
\begin{theorem}
\label{sumbounds} The expected time $T$ of convergence of the 
batch learner algorithm is bounded as follows:
\begin{equation}
\label{trivest} \sum_{i=1}^n \frac{1}{1-p_i} \geq T \geq 
\max_{1\leq i \leq n} \frac{1}{1-p_i}.
\end{equation}
\end{theorem}
The leftmost term in equation (\ref{trivest}) has been studied at 
length in \cite{kr1}. We state a version of the results of 
\cite{kr1} below:
\begin{theorem}
\label{allstab} Let $S=\sum_{i=1}^n \frac{1}{1-p_i},$ where the 
$p_i$ are independently identically distributed random variables 
with values in $[0, 1]$, with probability density $f$, such that 
$f(1-x) = x^\beta + O(x^{\beta - \delta}),\quad \delta > 0$ for 
$x\rightarrow 0$. Then If $\beta > 0$, then there exists a mean 
$m$, such that $\lim_{n \rightarrow \infty} \mathbb{P}(|S/n - m| 
> \epsilon) = 0,$ for any $\epsilon > 0.$ If $\beta = 0$, then 
$\lim_{n \rightarrow \infty} \mathbb{P}(|S/(n\log n) - 1| 
> \epsilon) = 0).$ Finally, if 
$-1 \leq \beta < 0,$ then $\lim_{n \rightarrow \infty} 
\mathbb{P}(S/n^{1/{\beta+1}} - C
> a) = g(a),$ where $\lim_{a \rightarrow \infty} g(a)= 0,$ and $C$ is 
an arbitrary (but fixed) constant, and likewise 
$$\mathbb{P}(S/n^{1/(\beta + 1)} < b) = h(b),$$ where $\lim_{a \rightarrow 0}h(a) = 0,$
\end{theorem}
The right hand side of Eq. (\ref{trivest}) is easier to 
understand. Indeed, let $p_1, \dots, p_n$ be distributed as usual 
(and as in the statement of Theorem \ref{allstab}. Then 
\begin{theorem}\label{expmin}
The expected value of $\max_{1 \leq i \leq n} p_i$ equals $1 - C 
n^{-1/{1+\beta}},$ for some positive constant $C$.
\end{theorem}
\begin{proof}
First, we change variables to $q_i = 1 - p_i$. Obviously, the 
statement of the Theorem is equivalent to the statement that $E = 
\mathbf{E}(\min_{1 \leq i \leq n} q_i) = C  n^{-1/{1+\beta}}$. We 
also write $h(x) = f(1-x),$ and similarly for the primitives $H$ 
and $F$. Now, the probability of that all of the $q_i$ are 
greater than some fixed $y$ equals $1-(1-H(y))^n,$ so that 
$$E = \int_0^1 t d\left[1-(1-H(t))^n\right] = \int_0^1 (1-H(t))^n d t.$$
Perform the change of variables $t = u/n^{1/(1+\beta)}$, to get 
\begin{equation}
\label{firstint} E = \frac{1}{n^{1+\beta}} 
\int_0^{n^{1/{1+\beta}}} (1-H(u/n^{1/(1+\beta)}))^n du. 
\end{equation}
For $u \ll n^{1/(1+\beta)}$, we can write $H(u/n^{1/(1+\beta)} 
\asymp u^{\beta + 1}/n H^\prime,$ where $H^\prime$ is a constant. 
We also know that $H$ is a monotonic function so if we break up 
the integral above as 
\begin{equation}
\label{secondint} E = \frac{1}{n^{1/(1+\beta)}} 
\left[\int_0^{n^{1/(2 (1 + \beta))}} + \int_{n^{1/(2 (1 + 
\beta))}}^{n^{1/(1 + \beta)}}\right] (1-H(u/n^{1/(1+\beta)}))^n 
du,
\end{equation}
we see that the first integral approaches $C = \int_0^\infty 
\exp(-u^{1/(1+\beta)}) d u,$ while the second integral goes to 0. 
Note that the proof also evaluates $C$.
\end{proof}
We need one final observation:
\begin{theorem}
The variable $n^{1/(1+\beta)} \min_{i=1}^n q_i$ has a limiting 
distribution with distribution function $G(x) = 
1-\exp(-x^{1+\beta}).$
\end{theorem}
\begin{proof}
Immediate from the proof of Theorem \ref{expmin}.
\end{proof}

We can now put together all of the above results as follows.
\begin{theorem}
\label{allgen}
 Let $p_1, \dots, p_k$ be independently distributed 
with common density function $f$, such that $f(1-x) = c x^\beta + 
O(x^{\beta + \delta}),$ $\delta > 0$. Let $T$ be the expected 
time of the convergence of the batch learning algorithm with 
overlaps $p_1, \dots, p_k$. Then, if $\beta > 0$, then there 
exist $C_1, C_2$, such that  $C_1 n^{1/(1+\beta)} \leq T \leq C_2 
n$, with probability tending to $1$ as $n$ tends to $\infty$. If 
$\beta = 0$, then there exist $C_1, C_2$, such that $C_1 n \leq T 
\leq C_2 n \log n$, with probability tending to one as $n$ tends 
to $\infty.$ If $\beta > 0$, then $C^{-1} n^{1/(\beta + 1)} \leq 
T \leq C n^{1/(\beta + 1)}$ with probability tending to $0$ as 
$C$ goes to infinity.
\end{theorem}

The reader will remark that in the case that $\beta > 0$, the 
upper and lower bounds have the same order of magnitude as 
functions of $n$.

\section{Independent concepts}
independence hypothesis, whereby an application of the 
inclusion-exclusion principle gives us:

\thm{lemma}{\label{latmost} The probability $l_k$ that we have 
 learned the concept $R_0$ after $k$ steps is given by 
$$
l_k=\prod_{i=1}^n(1-p_i^k).
$$
}

Note that the probability $s_k$ of winning the game \emph{on the 
$k$-th step} is given by $s_k = l_k - l_{k-1}= (1-l_{k-1}) - 
(1-l_k)$. Since the expected number of steps $T$ to learn the 
concept is given by
$$T = \sum_{k=1}^\infty k s_k,$$
we immediately have  $$T = \sum_{k=1}^\infty (1-l_k)$$
\thm{lemma}{\label{letime} The expected time $T$ of learning the 
concept $R_0$ is given by
$$
T = \sum_{k=1}^\infty \left(1-\prod_{i=1}^n 
\left(1-p_i^k\right)\right).
$$
} 
Since the sum above is absolutely convergent, we can expand the 
products and interchange the order of summation to get the 
following formula for $T$:

\begin{equation}
\label{subsum} T = \sum_{s\subseteq \{1, \dots, n\}} (-1)^{|s|-1} 
\sum_{k=1}^\infty p_s^k = \sum_{s\subseteq \{1, \dots, n\}} 
(-1)^{|s|-1} \left(\frac{1}{1-p_s} - 1\right),
\end{equation}
where we have identified subsets of $\{1, \dots, n\}$ with the 
corresponding multindexes.

The formula \ref{subsum} is useful in and of itself, but we now 
use it to attempt to get the expectation of the expected time of 
success $T$ under our distribution and independence assumption. 
For this we shall need the following:
\thm{definition}{\label{zdef} Let $\mathcal{F}$ be a probability 
distribution on an interval $I$, and let $m_k(\mathcal{F}) = 
\int_I x^k\mathcal{F}(d x)$ be the $k$-th moment of 
$\mathcal{F}$. Then the \emph{moment zeta function of 
$\mathcal{F}$} is defined to be 
$$\zeta_{\mathcal{F}}(s) = \sum_{k=1}^\infty m_k^s(\mathcal{F}),$$ whenever the sum is defined.
}
\thm{lemma}{\label{zetalemma} Let $\mathcal{F}$ be a probability 
distribution as above, and let $x_1, \dots, x_n$ be independent 
random variables with common distribution $\mathcal{F}$. Then
\begin{equation}
\mathbb{E}\left(\frac{1}{1-x_1 \dots x_n}\right) = 
\zeta_{\mathcal{F}}(n).
\end{equation}
In particular, the expectation is undefined whenever the zeta 
function is undefined. }
\begin{proof}
Expand the fraction in a geometric series and apply Fubini's 
theorem.
\end{proof}
\thm{example} { For $\mathcal{F}$ the uniform distribution on 
$[0, 1]$, $\zeta_{\mathcal{F}}$ is the familiar Riemann zeta 
function. Notice that this is \emph{not} defined for $n=1$ -- 
this will be important in the sequel.}

It should be noted that in the case we are interested in 
(distributions supported in $[0, 1]$), the asymptotics of the 
moments are determined by the local properties of the 
distribution at $1$, up to exponentially decreasing error terms. 
So, if $f(1-x) \asymp x^\beta$ (recall that $f$ is the density), 
we see that the $k$-th moment of $\mathcal{F}$ is asymptotic to 
$C k^{-(1+\alpha)},$ for some constant $C$.  To show this, we 
first define the \emph{Mellin transform} of $f$ to be 
$$\mathcal{M}(f)(s) = \int_0^1 f(x) x^{s-1} d x.$$ We see that 
$m_k(\mathcal{F}) = \mathcal{M}(f)(k+1).$ Mellin transform is 
very closely related to the Laplace transform. Indeed, making the 
substitution $x = \exp(-u)$, we see that $$\mathcal{M}(f) = 
\int_0^\infty f(\exp(-u)) \exp(-s u) d u,$$ so the Mellin 
transform of $f$ is equal to the Laplace transform of $f \circ 
\exp.$ Now, the asymptotics of the Laplace transform are easily 
computed by Laplace's method, and in the case we are interested 
in, Watson's lemma (see, eg, \cite{benorsz}) tells us that if 
$f(x) \asymp c (1-x)^\beta$, then $\mathcal{M}(f)(s) \asymp c 
\Gamma(\beta) x^{-(\beta + 1)}.$ In particular, 
$\zeta_{\mathcal{F}}(s)$ is defined for $s 
>1/(1+\beta)$. Below we shall analyze three cases (though the 
analysis is almost the same in the three cases, there are some 
important variations). In the sequel, we set $\alpha = \beta + 1$.
\section{$\alpha > 1$}
\label{isdef} 
In this case, we use our assumptions to rewrite Eq. 
(\ref{subsum}) as 
\begin{equation}
\label{subsum2} 
T = - \sum_{k=1}^n \binom{n}{k}(-1)^k \zeta_{\mathcal{F}}(k).
\end{equation}
This, in turn, can be rewritten (by expanding the definition of 
zeta) as
\begin{equation}
\label{subsum3} T = - \sum_{j=1}^\infty 
\left[\left(1-m_j(\mathcal{F})\right)^n-1\right]
\end{equation}
Since the term in the sum is monotonically decreasing, the sum in 
Eq. (\ref{subsum3}) can be approximated by an integral (of 
\emph{any} monotonic interpolation $m$ of the sequence 
$m_j(\mathcal{F})$; however there is no reason not to set $m(x) = 
\mathcal{M}(f)(x+1)$), with error bounded by the first term, 
which is, in term, bounded in absolute value by $2$, to get 
\begin{equation}
\label{approx1} T = - \int_1^\infty \left[(1-m(x))^n -1\right] d 
x + O(1),
\end{equation}
where the error term is bounded above by $2$.

Now, let us assume that $m(x)$ is of order $x^{-\alpha}$ for some 
$\alpha > 1$. We substitute $x = n^{1/alpha}/u$, to get
\begin{equation}
\begin{split}
 T &=  n^{1/\alpha}\int_0^{n^{1/\alpha}}
\frac{\left[1-(1-m(n^{1/\alpha}/u)^n \right]}{u^2} d u + O(1)\\ 
&= 
n^{1/\alpha}\int_0^{n^{1/\alpha}}\frac{\left[1-(1-m^\prime(u)u^\alpha/n)^n 
\right]}{u^2} d u + O(1)\\ & = 
n^{1/\alpha}\left(\int_0^{n^{1/2\alpha}} + 
\int_{n^{1/2\alpha}}^{n^{1/\alpha}}\right)
\frac{\left[1-(1-m^\prime(u)u^\alpha/n)^n \right]}{u^2} d u + 
O(1) ,
\end{split}
\end{equation}
where $m^\prime$ is a bounded (asymptotically constant) function. 
In the second integral the integrand is bounded above by $1/u^2$, 
so the contribution from that integral goes to $0$, while in the 
first integral we can approximate $(1-m^\prime u^\alpha/n)^n$ by  
$\exp(-m^\prime u^\alpha)$, and the contribution from that 
integral goes to 
\begin{equation}
\label{mainalpha} T = n^{1/\alpha} 
\int_0^\infty\frac{1-\exp(-m^\prime(u) u^\alpha)}{u^2} d u + O(1) 
\asymp C n^{1/\alpha}.
\end{equation}
\section{$\alpha = 1$}
\label{medalpha} In this case, $f(x) = c + o(1)$ as $x$ 
approaches $1$.  It is not hard to see that 
$\zeta_{\mathcal{F}}(n)$ is defined for $n \geq 2$. We break up 
the expression in Eq. (\ref{subsum}) as 
\begin{equation}
\label{subsumm} T = \sum_{j=1}^n {\frac{1}{1-p_j} - 1} + 
\sum_{s\subseteq \{1, \dots, n\}, \quad |s| > 1} 
 (-1)^{|s|-1} 
\left(\frac{1}{1-p_s} - 1\right).
\end{equation}
Let 
\begin{gather*} T_1 = \sum_{j=1}^n {\frac{1}{1-p_j} - 1},\\
 T_2 = \sum_{s\subseteq \{1, \dots, n\}, \quad |s| > 1} 
 (-1)^{|s|-1} 
\left(\frac{1}{1-p_s} - 1\right).
\end{gather*}
 The first sum $T_1$ has 
no expectation, however $T_1/n$  does have have a stable 
distribution centered on $c \log n + c_2$. We will keep this in 
mind, but now let us look at the second sum  $T_2$. It can be 
rewritten as 
\begin{equation}
\label{subsumm2} T_2 = - \sum_{j=1}^\infty 
\left[\left(1-m_j(\mathcal{F})\right)^n-1 + n m_j\right].
\end{equation}
The same method as in section \ref{isdef} under the assumption 
that the $k$-th moment is asymptotic to $k^\alpha$ (this time for 
$\alpha \leq 1$) can be used to write 
\begin{equation}
\begin{split}
T_2 &= n \int_0^n \frac{\left[1-n m(n/u) - (1-m(n/u)^n 
\right]}{u^2} d u + O(1)\\ &= n\left(\int_0^{n^{1/2}} + 
\int_{n^{1/2}}^n\right) \frac{\left[1- m^\prime(u) u - 
(1-m^\prime(u)u/n)^n \right]}{u^2} d u + O(1).
\end{split}
\end{equation} The conclusion differs somewhat from that of section \ref{isdef} in that  we get an 
additional term of $c n \log n$, where $c = \lim_{x \rightarrow 
1} f(x) = \lim_{j \rightarrow \infty} j m_j$. This term is equal 
(with opposing sign) to the center of the stable law satisfied by 
$T_1$, so in case $\alpha = 1$, we see that $T$ has no 
expectation but satisfies a \emph{law of large numbers}, of the 
following form: 
\begin{theorem}[Law of large numbers]
There exists a constant $C$ such that $\lim_{y \rightarrow 
\infty} \mathbf{P}(|T/n - C| > y) = 0.$
\end{theorem}
\section{$\alpha <1$}
\label{smallalpha} In this case the analysis goes through as in 
the preceding section when $\alpha > 1/2$, but then runs into 
considerable difficulties. However, in this case we note that 
Theorem \ref{allgen} actually gives us tight bounds. 
\section{The inevitable comparison}
We are now in a position to compare the performance of the batch 
learning algorithm with that of the memoryless learning algorithm 
and of learning with full memory, as summarized in Theorem 
\ref{mainprev}. We combine our computations above with the 
observation that the batch learner algorithm converges 
geometrically (Lemma \ref{latmost}), to get: 
\thm{theorem} {\label{batchthm} Let $N_\Delta$ be the number of 
steps it takes for the student (with probability $1$) to have 
probability $1 - \Delta$ of learning the concept using the batch 
learner algorithm. Then we have the following estimates for 
$N_\Delta$:
\begin{itemize}
\item
If the distribution of overlaps is \emph{uniform}, or more 
generally, the density function $f(1-x)$  at $0$ has the form 
$f(x) = c + O(x^\delta),$ $\delta, c > 0,$ then $N_\Delta=|\log 
\Delta|\Theta(n)$ 
\item 
If the probability density function $f(1-x)$ is asymptotic to 
$x^\beta + O(x^{\beta - \delta}), \quad \delta, \beta > 0$, as 
$x$ approaches $0$, then we have $N_\Delta=|\log 
\Delta|\Theta(n^{1/(1+\beta)})$; 
\item 
If the asymptotic behavior is as above, but $-1 < \beta < 0$, 
then $N_\Delta=|\log \Delta|\Theta(n^{1/(1+\beta)}).$ 
\end{itemize}}
Comparing Theorems \ref{mainprev} and \ref{batchthm}, we see that 
batch learning algorithm is uniformly superior for $\beta \geq 
0$, and the only one of the three to achieve \emph{sublinear} 
performance whenever $\beta 
> 0$ (the other two \emph{never} do better than linearly, unless 
the distribution $\mathcal{F}$ is supported away from $1.$) On 
the other hand, for $\beta < 0$, the batch learning algorithm 
performs comparably to the memoryless learner algorithm, and 
worse than learning with full memory.

\end{document}